\documentclass[10pt,journal,compsoc]{IEEEtran}

\ifCLASSOPTIONcompsoc
  \usepackage[nocompress]{cite}
\else
  \usepackage{cite}
\fi

\usepackage{xspace}
\usepackage[table,dvipsnames]{xcolor}

\usepackage{graphicx}
\usepackage{amsmath}
\usepackage{amssymb}
\usepackage{booktabs}
\usepackage{multirow}
\usepackage{pifont}%

\usepackage{siunitx}
\usepackage{setspace}
 \DeclareSIUnit\pixel{px}

\usepackage[format=plain,labelformat=simple,labelsep=period,font=small,compatibility=false]{caption}
\usepackage[font=footnotesize,skip=3pt,subrefformat=parens]{subcaption}

\usepackage{array}
\newcolumntype{L}[1]{>{\raggedright\let\newline\\\arraybackslash\hspace{0pt}}m{#1}}
\newcolumntype{C}[1]{>{\centering\let\newline\\\arraybackslash\hspace{0pt}}m{#1}}
\newcolumntype{R}[1]{>{\raggedleft\let\newline\\\arraybackslash\hspace{0pt}}m{#1}}

\usepackage[pagebackref=true,breaklinks=true,letterpaper=true,colorlinks,bookmarks=false]{hyperref}

\usepackage[capitalize]{cleveref}
\crefname{section}{Sec.}{Secs.}
\Crefname{section}{Section}{Sections}
\Crefname{table}{Table}{Tables}
\crefname{table}{Tab.}{Tabs.}
\usepackage{xr}
\newcommand{\cmark}{\ding{51}}%

\makeatletter
\DeclareRobustCommand\onedot{\futurelet\@let@token\@onedot}
\def\@onedot{\ifx\@let@token.\else.\null\fi\xspace}

\def\eg{\emph{e.g}\onedot} 
\def\ie{\emph{i.e}\onedot} 
\def\cf{\emph{c.f}\onedot}

\def\etal{\emph{et al}\onedot}
\makeatother

\newcommand{\change}[1]{#1}
\renewcommand*{\and}{\hspace{0.9cm}}

\newcommand{\myparagraph}[1]{\vspace{1pt}\noindent{\bf{#1}}}

\begin{document}

\title{PhenoBench: A Large Dataset and Benchmarks for Semantic Image Interpretation\\ in the Agricultural Domain}

\author{
  Jan Weyler,~
  Federico Magistri,~
  Elias Marks,~
  Yue Linn Chong,~
  Matteo Sodano,\\
  Gianmarco Roggiolani,~
  Nived Chebrolu,~
  Cyrill Stachniss,~and
  Jens Behley
  \IEEEcompsocitemizethanks{
    \IEEEcompsocthanksitem J. Weyler, F. Magistri, E. Marks, Y.L. Chong, M. Sodano, G. Roggiolani, and J. Behley are with the Center for Robotics, University of
    Bonn, Germany. 
    E-mails: \{firstname.lastname\}@igg.uni-bonn.de
    \IEEEcompsocthanksitem N. Chebrolu is with the University of Oxford, UK. \protect\\ E-mail: nived@robots.ox.ac.uk
    \IEEEcompsocthanksitem C. Stachniss is with the Center for Robotics, University of Bonn, Germany, the University of Oxford, UK, and the Lamarr Institute for Machine Learning and Artificial Intelligence, Germany. \protect\\ E-Mail: cyrill.stachniss@igg.uni-bonn.de}
}

\IEEEtitleabstractindextext{%
  \begin{abstract}
    The production of food, feed, fiber, and fuel is a key task of agriculture, which has to cope with many challenges in the upcoming decades, e.g., a higher demand, climate change, lack of workers, and the availability of arable land. Vision systems can support making better and more sustainable field management decisions, but also support the breeding of new crop varieties by allowing temporally dense and reproducible measurements. Recently, agricultural robotics got an increasing interest in the vision and robotics communities since it is a promising avenue for coping with the aforementioned lack of workers and enabling more sustainable production. While large datasets and benchmarks in other domains are readily available and enable significant progress, agricultural datasets and benchmarks are comparably rare. We present an annotated dataset and benchmarks for the semantic interpretation of real agricultural fields. Our dataset recorded with a UAV provides high-quality, pixel-wise annotations of crops and weeds, but also crop leaf instances at the same time. Furthermore, we provide benchmarks for various tasks on a hidden test set comprised of different fields: known fields covered by the training data and a completely unseen field. Our dataset, benchmarks, and code are available at \url{https://www.phenobench.org}.
  \end{abstract}
}

\maketitle

\IEEEdisplaynontitleabstractindextext
\IEEEpeerreviewmaketitle

\newlength{\teaserwidth}
\setlength{\teaserwidth}{0.24\textwidth}
\newlength{\teaserspace}
\setlength{\teaserspace}{0.05cm}

\begin{figure*}
  \includegraphics[width=\teaserwidth]{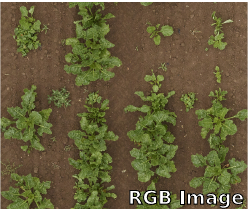}
  \hspace{\teaserspace}
  \includegraphics[width=\teaserwidth]{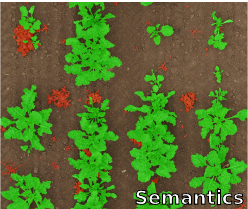}
  \hspace{\teaserspace}
  \includegraphics[width=\teaserwidth]{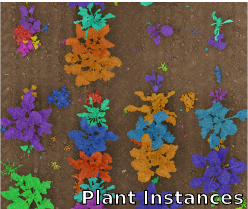}
  \hspace{\teaserspace}
  \includegraphics[width=\teaserwidth]{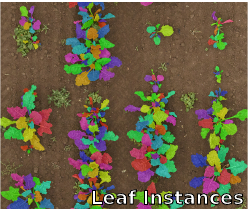}
  \caption{
    Our dataset, called \emph{PhenoBench}, provides dense semantic plant-level instance annotations~(shown by different colors) of sugar beet crops and weeds~(green and red in the semantics) and leaf-level instance annotations of crops~(different colors correspond to different instances) for high-resolution images recorded with a UAV. The dataset consists of images collected at different times  during a growing season, which captures various growth stages of plants.\label{fig:teaser} \vspace{-0.2cm}
  }
\end{figure*}

\ifCLASSOPTIONcompsoc
  \IEEEraisesectionheading{\section{Introduction}\label{sec:introduction}}
\else
  \section{Introduction}\label{sec:introduction}
\fi

\IEEEPARstart{T}{he} agricultural production of food, feed, fiber, and fuel has to cope with several challenges in the upcoming decades.
The world population is increasing, yet the availability of arable land is limited or even decreasing, climate change increased uncertainties in crop yield, and we observe substantial losses in biodiversity~\cite{duckett2018arxiv}.
At the same time, agricultural practices need to be more sustainable and have to reduce the use of agrochemical inputs, \ie, herbicides and fertilizers that potentially negatively impact yield~\cite{horrigan2002ehp} and the environment.

Robots and drones using vision-based perception systems could help with these challenges by offering tools to make better, more sustainable field management decisions and providing supporting tools for breeding new varieties of crops by estimating plant traits in a reproducible manner~\cite{pound2017gs}.
Such visual perception systems enable the development of agricultural robots that can support the monitoring of fields and replace labor-intensive tasks such as manual weeding~\cite{walter2018icpa}.
Additionally, they potentially enable more targeted crop management, where agrochemicals are applied precisely and only where needed, thereby reducing the negative effects on the environment~\cite{linaza2021agronomy, storm2024eja}.

With the advent of deep learning for visual perception~\cite{lecun2015nature, krizhevsky2012cacm}, the field of computer vision has made tremendous progress in image interpretation, achieving remarkable results in several domains.
Datasets and associated benchmarks~\cite{cordts2016cvpr,lin2014eccv,neuhold2017iccv} were essential for achieving this progress as they provide a testbed for developing novel algorithms but also provided the necessary data to tackle novel tasks.
Progress can be tracked quantitatively with metrics that measure the performance of developed approaches against benchmarks using hidden test sets.
Novel tasks with increasing complexity drive the progress of the field by posing novel challenges for the community.

In this paper, we aim to provide a large dataset together with benchmarks for semantic interpretation under real-field conditions enabling similar progress in the agricultural domain.
We target multiple tasks: semantic segmentation, panoptic segmentation, plant detection, leaf detection, and the novel task of hierarchical panoptic segmentation %
\change{that provides a coarse-to-fine interpretation of plants.}

For this purpose, we recorded high-resolution images with unmanned aerial vehicles~(UAV) of sugar beet fields under natural lighting conditions over multiple days, capturing a large range of growth stages. We annotated these images with dense, pixel-wise annotations to identify sugar beet crops and weeds at an instance level, as needed for semantic segmentation and plant-level instance segmentation tasks. Additionally, we labeled leaf instances of crops to enable the investigation of leaf instance segmentation (see \cref{fig:teaser}). Furthermore, we provide temporal association of plant instances over the different dates, which allows to identify individual plants at different growth stages.

The combination of plant-level and leaf-level annotations enable the investigation of novel tasks %
needed for a holistic semantic interpretation in the agricultural domain.
One such task is the hierarchical panoptic segmentation %
\change{that targets to} segment individual leaves and assign them to their associated plant instance to predict the total number of leaves per plant.
Plant scientists and breeders commonly assess this information to describe the growth stage of individual plants, which is also linked to yield potential and plant performance~\cite{lancashire1991aab}.
However, this in-field assessment is \change{nowadays} done manually outside greenhouses, which is laborious and time-consuming~\cite{minervini2015spm}.
Thus, developing vision systems to assess these properties per plant automatically is essential for large-scale, sustainable crop production.

Our provided data shows distinct challenges in terms of plant variation and overlap between different plant and leaf instances that are distinct in the agricultural domain.
Such challenges are seldomly addressed by general segmentation approaches prevalent in man-made environments, as shown by our experimental results, where we challenged several state-of-the-art approaches but also provide results for more domain-specific approaches for the agricultural domain.

In summary, our main contributions are:
\begin{itemize}
  \item We present a large dataset for plant segmentation providing accurate instance annotations at the level of plants and leaves.
  \item We provide benchmark tasks on a hidden test set for evaluating semantic, instance, and panoptic segmentation, and detection approaches targeted at \change{plants enabling} reproducible and unbiased evaluation \change{of novel plant perception approaches}.
  \item We provide baseline results for general and domain-specific models for plant and leaf detection, but also semantic, instance, and panoptic segmentation.
\end{itemize}

We believe that the effort in generating high-quality annotations and establishing reliable benchmarks for multiple tasks %
\change{with} a hidden test set will accelerate progress in semantic perception of agricultural fields and potentially lead to novel avenues of research in this important domain. We make our dataset and benchmarks\footnote{\url{https://www.phenobench.org}}, code for visualizing predictions and computing metrics\footnote{\url{https://github.com/PRBonn/phenobench}}, and baselines\footnote{\url{https://github.com/PRBonn/phenobench-baselines}} with code, \change{checkpoints}, and predictions publicly available.

\section{Related Work}

In recent years, dense, pixel-wise semantic interpretation of images, \ie, semantic, instance, and panoptic segmentation~\cite{kirillov2019cvpr-ps}, made rapid progress due to advances in deep learning~\cite{lecun2015nature}, but also thanks to the availability of large-scale datasets for object detection~\cite{lin2014eccv,everingham2010ijcv,everingham2015ijcv}, semantic segmentation~\cite{cordts2016cvpr,neuhold2017iccv}, instance segmentation~\cite{lin2014eccv}, and lately panoptic segmentation~\cite{lin2014eccv,cordts2016cvpr,neuhold2017iccv}.

\begin{table*}[t]
  \centering
  \tabcolsep8pt
  \small{
    \begin{tabular}{lcccC{0.9cm}C{1.0cm}C{0.9cm}C{0.9cm}C{0.9cm}cC{1.2cm}}
      \toprule
      \multirow{2}{*}{Dataset}                                   & \multirow{2}{*}{\#Images} & \multirow{2}{*}{Image Size} & \multicolumn{3}{c}{Crop} & \multicolumn{2}{c}{Weed} & \multirow{2}{*}{Field?} & \multirow{2}{1.2cm}{Hidden Test Set?}                            \\
      \cmidrule(lr){4-6}\cmidrule(lr){7-8}                       &                           &                             & Sem.                     & Inst.                    & Leaves                  & Sem.                                  & Inst.  &        &        \\
      \midrule
      CWFID~\cite{haug2014eccv}                                  & 60                        & 1291 $\times$ 966           & \cmark                   &                          &                         & \cmark                                &        & \cmark &        \\
      CVPPP~\cite{bell2016zenodo,minervini2016prl} & 1,311                     & 2048 $\times$ 2448$^1$      &                          &                          & \cmark                  &                                       &        &        & \cmark \\
      Carrot-Weed~\cite{lameski2017icti}                         & 39                        & 3264 $\times$ 2448          & \cmark                   &                          &                         & \cmark                                &        & \cmark &        \\
      Sugar beets~\cite{chebrolu2017ijrr}                        & 280                       & 1296 $\times$ 966           & \cmark                   &                          &                         & \cmark                                &        & \cmark &        \\
      WeedMap~\cite{sa2018rs}                                    & 1,670                     & 480 $\times$ 360            & \cmark                   &                          &                         & \cmark                                &        & \cmark &        \\
      Carrots-Onion~\cite{bosilj2019jfr}                         & 40                        & 2464 $\times$ 2056          & \cmark                   &                          &                         & \cmark                                &        & \cmark &        \\
      Oil Radish~\cite{mortensen2019cvpr}                        & 129                       & 1600 $\times$ 1600          & \cmark                   &                          &                         & \cmark                                &        & \cmark &        \\
      Sunflower~\cite{fawakherji2021ras}                         & 500                       & 1296 $\times$ 966           & \cmark                   &                          &                         & \cmark                                &        & \cmark &        \\
      GrowliFlowers~\cite{kierdorf2022jfr}                       & 2,198                     & 448 $\times$ 368            & \cmark                   & \cmark                   & \cmark                  &                                       &        & \cmark &        \\
      CropAndWeed~\cite{steininger2023wacv}                      & 8,034                     & 1920 $\times$ 1088          & \cmark                   &                          &                         & \cmark                                &        & \cmark &        \\
      \rowcolor{SpringGreen}
      PhenoBench (Ours)                                          & 2,872                     & 1024 $\times$ 1024          & \cmark                   & \cmark                   & \cmark                  & \cmark                                & \cmark & \cmark & \cmark \\
      \bottomrule
    \end{tabular}
  }
  \caption{Comparison of datasets in the agricultural domain providing \emph{dense pixel-wise annotations}.
    For the crop and weed, we indicate if semantic segmentation (Sem.), plant instances (Inst.), and leaf instances (Leaves) are densely annotated.
    We also record if the dataset was recorded under field conditions, as opposed to under lab conditions (Field?). Furthermore, we note if there is a hidden test set, such that approaches do not have access to test set labels (Hidden Test Set?).\quad \small{$^1$We report maximum image size, as it ranges from $\SI{441}{\pixel}\times\SI{441}{\pixel}$ to $\SI{2048}{\pixel}\times\SI{2448}{\pixel}$.} \vspace{-0.2cm}}
  \label{tab:plant_datasets}

\end{table*}

Despite the availability of large datasets in man-made environments, the agricultural domain faces different challenges, such as large intra-class variability due to plant growth.
Thus, there has been interest in large datasets to enable studying perception in the agricultural domain~\cite{lu2020cea}.
However, accurately dense annotated and large agricultural datasets in combination with reproducible benchmarks on a hidden test set are still missing today, see \cref{tab:plant_datasets}.

In particular, the crop/weed field image dataset~\text{(CWFID)} by Haug \etal~\cite{haug2014eccv} is one of the first semantic segmentation datasets that provides pixel-level annotations of semantics for plants, \ie, sugar beets and weeds using a multispectral camera.
Lameski \etal~\cite{lameski2017icti} also provides a dataset for crops, \ie, carrots and weed segmentation.
CVPPP~\cite{minervini2016prl, bell2016zenodo} is one of the first datasets providing annotations for leaves in images of individual tobacco and arabidopsis plants recorded in a lab environment,
which is also the basis for a series of workshops and competitions hosted at CVPR and ICCV.
The dataset by Chebrolu \etal~\cite{chebrolu2017ijrr} provides images of sugar beets and weeds recorded by a ground robot under real field conditions with a ground sampling distance~(GSD) of~$\SI{0.3}{\mm\per\pixel}$ and provides annotations for semantic segmentation.
Similar to our dataset, the WeedMap dataset~\cite{sa2018rs} provides imagery of UAVs covering a large field with sugar beets and weeds.
In contrast to our dataset, where we provide the original camera data, WeedMap first generated orthophotos via bundle adjustment.
While we considered this option, we noticed that the lack of a detailed elevation model usually leads to artifacts on the boundaries of the plants.
Additionally, the images of WeedMap have a coarse GSD between~$\SI{8.2}{\cm\per\pixel}$ and~$\SI{13}{\cm\per\pixel}$ while our images have a GSD of~$\SI{1}{\mm\per\pixel}$ to assess detailed information for individual plants.
The Sunflowers dataset~\cite{fawakherji2021ras} provides images collected with a multi-spectral sensor providing RGB and near-infrared images from a ground robot.
The {Agriculture-Vision} dataset~\cite{chiu2020cvpr} contains aerial images with a coarse GSD between~$\SI{10}{\cm\per\pixel}$ and~$\SI{20}{\cm\per\pixel}$ with corresponding
annotations that covers rather large areas but not individual plants, \eg, regions with nutrient deficiencies and weed clusters.
More recently, the GrowliFlowers dataset~\cite{kierdorf2022jfr} provides images
recorded with a UAV showing multiple growth stages of cauliflowers.
\change{While we recorded images on three dates roughly a week apart, this dataset
  contains images captured on four different dates, also roughly a week apart.
  Therefore, it captures an extended period of one month.}

Lately, the CropAndWeed dataset~\cite{steininger2023wacv} provides RGB images taken close to the field canopy showing a large variety of crops and weeds.
While the number of annotated images is large, the pixel-wise annotations have been semi-automatically annotated exploiting a pre-segmentation via a deep neural network to lower the annotation effort.
However, this sometimes leads to incomplete annotations and notable annotation artifacts.
Also in our experience, we noted that correcting annotations is quite tedious and can counter-intuitively lead to even larger annotation effort as boundary regions generated using contemporary segmentation approaches almost always need to be corrected, which is also the part that takes most of the annotation time.

\change{The recently published RumexLeaves dataset~\cite{gueldenring2023ral} provides fine-grained annotations of leaves of the Rumex obtusifolius L., which is a problematic weed in grasslands. Besides the leaf annotations of this particular plant, the dataset also provides more fine-grained vein and stem annotations that allows to get insights into the plant physiology corresponding to traits relevant for plant phenotyping.}

Besides the aforementioned closely related datasets that also provide dense pixel-wise annotations, there have been recently also several datasets in the agricultural domain released for wheat detection~\cite{david2021arxiv}, localization and mapping~\cite{pire2019ijrr, imperoli2018ral}, image classification of weed species~\cite{olsen2019sr}, detection for phenotyping~\cite{madsen2020rs}, crop row detection~\cite{winterhalter2018ral}, or fruit detection~\cite{sa2016sensors}.
Additionally, there are a small number of available datasets for semantic interpretation of 3D agricultural data~\cite{schunck2021plosone,khanna2019plantmethods,dutagaci2020plantmethods}.

While recent interactive labeling approaches, like SegmentAnything~\cite{kirillov2023iccv}, can certainly speed-up labeling of instance masks with only weak annotations delivering compelling results, we target to generate a reliable and high-quality dataset and corresponding benchmark. Therefore, we resort to manual annotations from scratch, which entailed a rigorous correction and verification procedure to ensure accurate and consistent segmentation masks.

In contrast to the aforementioned datasets, which are great starting points for research, our dataset shows an unique level of annotations, including semantic and instance masks for crops and weeds of an overall larger number individual plants~(see \cref{tab:plant_datasets}). Furthermore, we provide temporally consistent instance ids of crops that allow to identify individual plants over multiple dates.
Note that our dataset provides large images with multiple completely visible plants, which is not always the case for other pixel-wise annotated datasets~\cite{kierdorf2022jfr, minervini2016prl}. Lastly, we enable comparable and reproducible results with the provided benchmarks on a hidden test set, \ie, labels are not released and the predictions are evaluated on a server via CodaLab~\cite{pavao2023jmlr}.

\section{Our Dataset}

In this section, we present our setup for data collection, explain the labeling process, and provide statistics to show the variability of the data.

\subsection{Data Collection}

Our dataset provides RGB images in real field conditions recorded by an UAV
equipped with a high-resolution camera that captures
imagery of the field. For recording the data, we employed a DJI M600 and used
the PhaseOne iXM-100 camera with a $\SI{80}{\mm}$ RSM prime lens mounted on a
gimbal to obtain {motion-stabilized} RGB images at a resolution of
${\SI{11664}{\pixel} \times \SI{8750}{\pixel}}$. The UAV was flying at a
height of approx.~$\SI{21}{\metre}$, resulting in a GSD of
$\SI{1}{\mm\per\pixel}$. For covering the entire field, we use the DJI Ground
Station Pro app to plan a flight that covers the field row-wise. We set the
forward overlap between consecutive images by motion vector at
$\SI{75}{\percent}$ and the side overlap between images placed in neighboring
rows at $\SI{50}{\percent}$. Each image is geo-referenced by using the on-board
GNSS.

\begin{figure}[t]
  \centering
  \includegraphics[width=\linewidth]{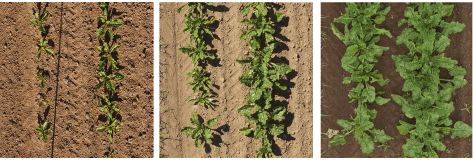}
  \caption{Variability in overlap and illumination of plants at the same part of the field on different recording dates.
    Theses examples show the variation in growth stages ranging from 4 leaf stage (early growth stage) to plants with over 20 leaves (later growth stage)
    and the variety of illuminations with sunny (left) and overcast (right) weather conditions.}
  \label{fig:growth_stages}
\end{figure}

We performed three missions roughly a week apart to capture different
growth stages of the plants. More specifically, we performed the flights on May
15, May 26, and June 6 in 2020.
Additionally, we used the same sensor setup to record images at four different points in
time in 2021 on a different field: May 20, May 28, June 1, and June 10.
As the data was captured in the open
field, we have a variety of different lighting conditions with sunny and also
overcast weather, as shown in \cref{fig:growth_stages}, which significantly changes the visual appearance of the plants.

\begin{figure}[t]
  \centering
  \includegraphics[width=\linewidth]{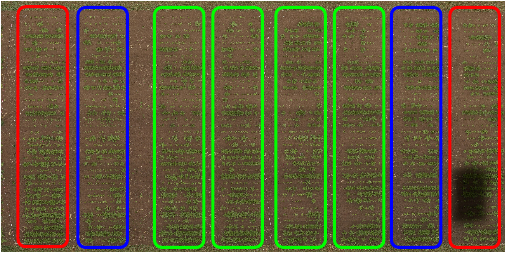}
  \caption{Orthophoto of the field recorded in 2020 and our spatial separation into rows for training (green), validation (blue), and testing (red). Due to the geo-referencing of the images, we extracted the same rows on each of the dates.}
  \label{fig:row_structure}
\end{figure}

From the approximately $\SI{1 300}{\meter\squared}$ sugar beets field located at
the Campus Klein-Altendorf farm between Meckenheim and Rheinbach, Germany
(50$^\circ$37'.51N, 6$^\circ$59'.32E), we selected eight crop rows that were
covered by the recording mission. To have a clear spatial separation between the
train and test set, we selected four crop rows for extracting training images,
two crop rows for validation, and two crop rows for testing purposes as shown in
\cref{fig:row_structure}. Additional data recorded in 2021 is only included in
the test set to evaluate also the performance in a setting of an unseen field
with the same crop but potentially different weeds.

\change{Specifically, the sugar beet field contains a mixture of two different
  crop varieties, \ie,~\mbox{BTS 440} and \mbox{Celesta KWS} that are both from
  distinct {agro-seed} companies and differ in their properties regarding a beet's
  mass and sugar yield. Furthermore, we observe six weed varieties that are most
  prominent in the field, \ie, Chenopodium album, Polygonum aviculare, Thlaspi
  arvense, Persicaria lapathifolia, Bilderdykia convolvulus, and Polygonum
  hydropiper.}

  \begin{figure}[t]
    \centering
    \includegraphics[width=\linewidth]{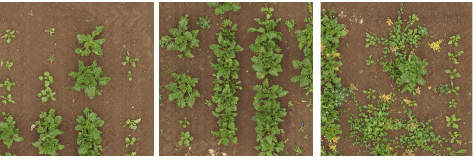}
    \caption{\change{Varying conditions of the field recorded at different locations,
        which are treated with different amounts of herbicides.
        From left to right: \mbox{Fully-herbicided}, \mbox{partially-herbicided},
        and \mbox{non-herbicided} field conditions recorded at the same day.}}
    \label{fig:fhphnh}
  \end{figure}

  \begin{figure}[t]
    \centering
    \includegraphics[width=0.7\linewidth]{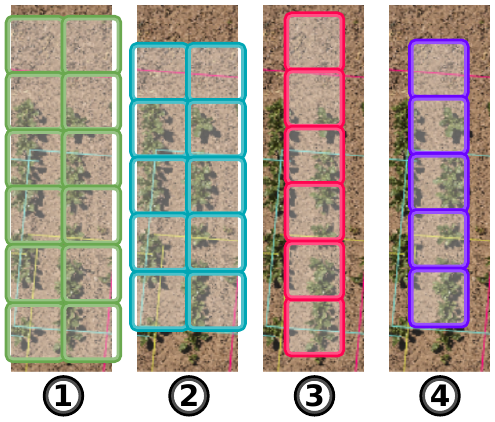}
    \caption{Extracted tiles \change{per} iteration such that \change{a row is densely} covered with tiles to ensure that all plants are completely visible in at least \change{one} tile. Annotations of tiles are transferred between iterations and aggregated in the global image $I_g$.}
    \label{fig:row_tiling}
  \end{figure}

The field belongs to a farm of the University of Bonn located at the Campus Klein-Altendorf. This allows us to conduct field studies and to study perception systems under varying conditions with respect to application of herbicides, which leads to different scenarios with fully~(conventional), partial~($80\%$ herbicides), and non-herbicided field conditions, \change{as shown in~\cref{fig:fhphnh}}.
In conventional farming and field management operations, such conditions with less or no herbicides are usually not observable.
While keeping most of the other field parameters constant, this makes our field setup \change{distinct} to other larger datasets, such as GrowliFlowers~\cite{kierdorf2022jfr} that recorded data only under conventional field management conditions with only a very few weeds.

\subsection{Labeling Process}

The full-sized images, which we denote as global images, $I_g$, are challenging to annotate due to their large size of ${\SI{11664}{\pixel} \times \SI{8750}{\pixel}}$.
To parallelize the labeling process and ensure no plant is missed,
we extracted from $I_g$ overlapping patches, $I_p$, of size ${\SI{2000}{\pixel} \times \SI{2000}{\pixel}}$.
We extracted multiple iterations of overlapping patches such that we always have in one of the resulting four tilings complete plants visible, \cf \cref{fig:row_tiling}.
As we ensure that each plant is fully visible in at least one of the patches, we
instructed our annotators to label only completely visible plants in $I_p$.

For labeling the plants and leaves at the same time, we developed a novel tool to enable a hierarchical annotation of the images.
Please see the supplement for a more detailed description of the labeling tool and the provided features.

We first labeled the plant instances of sugar beet crops and weeds, which was completed by 9 annotators investing a total of 800\,h.
Each iteration was validated and corrected before we transferred the annotations to the global images $I_g$.
Then, the next iteration is started with the transferred labels copied to the respective patches $I_p$, and these steps were repeated till the final fourth iteration.

Annotation of a single patch~$I_p$ ranged from approx. 1\,h for earlier growth stages to 3.5\,h for later growth stages where plants had significant overlap. In sum, we annotated 705 patches over all dates and crop rows.

\begin{table}[t]
  \centering
  \small{
    \begin{tabular}{L{1.55cm}C{1.3cm}C{1.3cm}C{1.3cm}C{1.3cm}}
      \toprule
      Split      & \#imgs    & \#crops  & \#weeds & \#leaves \\
      \midrule
      Train      & $1,407$   & $11,875$ & $8,141$ & $71,264$ \\
      Validation & $772$     & $6,482$  & $3,926$ & $35,503$ \\
      Test       & $693$     & $6,201$  & $4,291$ & $33,935$ \\
      \midrule
      Unlabeled  & $129,000$ & --       & --      & --       \\
      \bottomrule
    \end{tabular}}

  \caption{Dataset statistics of the provided splits. Note that we have a hidden test set, \ie, we have a server-sided evaluation~\cite{pavao2023jmlr}. We additional provide unlabeled data of the fields to enable studying of self-supervised pre-training.}
  \label{tab:statistics}
\end{table}

After the plant instances were labeled, we had 5 annotators labeling leaf instances.
Annotators were tasked with identifying crop leaves and annotation of a patch $I_p$ took approx.~1\,h to~2\,h depending on the number of visible crops.
With the masking of plant instances provided by our annotation tool, we ensure that we have consistent leaf labels that are inside the crop instance.
Thus, it is possible to associate each leaf instance with its corresponding crop based on the plant instance annotations.

To ensure high-quality, accurate annotations of plants and leaves, we furthermore had an additional round of corrections performed by four additional annotators that revised the annotations. More details on our quality assurance process is provided in the supplementary material.

In total, we had 14 annotators who invested 1,400\,h of annotation work and roughly 600\,h invested into additional validation and refinement, leading to an overall labeling effort of approximately 2,000\,h.

\subsection{Temporal Alignment}

As we recorded images in the same geographical location, we can furthermore provide temporally aligned plant instances, which enables the study of individual plant growth. By matching the occurrences of the same plants in different recordings we ensure that each crop plant has a unique instance id throughout our whole dataset.

To this end, we exploit the positions delivered by the RTK GNSS of the drone as initial guesses for a bundle adjustment procedure to determine the pose of the camera for each captured image in a global reference frame.
This allows us to project the crop center locations, computed as the centroid of the plant pixels, of plants appearing in all images of a mission into a common plane.

As the estimated poses of the camera are not completely free of noise, we use Hungarian matching~\cite{kuhn1955nrlq} based on the distances of crop centers to robustly associate instances of the same plant appearing in different images. To account for new crop instances but also missing crop instances, we only associate crop centers, when their distance is below a threshold of $15\,$cm, which was determined empirically.
We experimented with using GNSS poses to associate crop instances between different missions collected at different points in time, but found the inaccuracies of the localization to be too high for our purpose. We, therefore, manually associated around 10 plants between the different missions and used these datapoints to compute a transformation between each mission using a least squares approach.
Given those transformations we then associated the crop ids again by projecting them onto a common plane and matching them by the Hungarian algorithm.
Finally, we validated the temporal alignment by visualizing the matches between missions at different points in time.

\subsection{Dataset Statistics}

We finally extracted from the global images $I_g$ smaller images of size ${\SI{1024}{\pixel} \times \SI{1024}{\pixel}}$ to ensure that we have images containing complete crops at later growth stages, but also provide context such as the crop row structure. More specifically, we use the an overlap of $50$\% between extracted patches to ensure that plants in later growth stages are at least 50\% visible in the extracted patches.

\cref{tab:statistics} shows an overview of the number of extracted images for the different splits from the earlier described train/validation/test rows, the number of crop instances, the number of crop leaves, and the number of weed instances annotated. Note that only the test data includes data from 2020 and 2021.
As we ensured that we have completely annotated plants,
we are able to generate a visibility map and differentiate between mostly visible plants with at least 50\,\% visible pixels and partially visible plants.
Note that we provide a rather large validation set to allow researchers to conduct conclusive ablations studies.

In addition to the labeled data, we also provide unlabeled data from all fields, which can be exploited for pre-training, semi-supervised, or unsupervised domain adaptation, which we see as promising future avenue of research.

\begin{figure}[t]
  \centering
  \includegraphics[width=\linewidth]{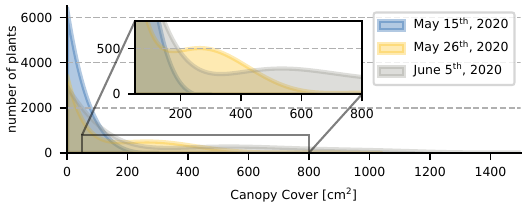}
  \caption{Distribution of crop sizes in terms of canopy cover in~cm$^2$ for mostly visible plants in the training and validation set.}
  \label{fig:plant_size_statistics}
\end{figure}

As motivated earlier, we recorded images under real-world conditions of real agricultural fields leading to a diverse range of plant appearances due to varying growth stages.
The crops are affected by different soil conditions leading to a variety of growth stages even on images of the same date.
This intra-class variability of the crops poses an interesting challenge for learning approaches that have to correctly segment or detect small but also large crops at the same time. The extra data from a different field captured in 2021 leads to even greater diversity of recording conditions, which is a common challenge in the agricultural domain.

Additionally, we observe a large variability in terms of overlap between plants. They are clearly separated at the beginning of the recording campaign but show a considerable overlap at the last recording date.
\cref{fig:growth_stages} shows the same area of the field over the course of three weeks showing the variation in terms of growth stage but also the overlap between crops.

In \cref{fig:plant_size_statistics}, we provide an overview of the plant sizes \change{per data collection day} in terms of the area covered by the plant instances that shows the diversity in terms of growth stages.
\change{While on May 20$^\text{th}$ plants with a small coverage are predominately present, the plant area of plants naturally increased in the following weeks.
  On May 26$^\text{th}$, the amount of larger plants increases. At the latest date, June 5$^\text{th}$, the amount of larger plants further increases and the distribution gets more long-tailed as now all plants directly compete for space, which is also visually visible from the larger overlap between neighboring plants. Thus, only few plants are able to develop a larger canopy cover.}

\begin{figure}[t]
    \centering
    \vspace{0.3cm}
    \includegraphics[width=\linewidth]{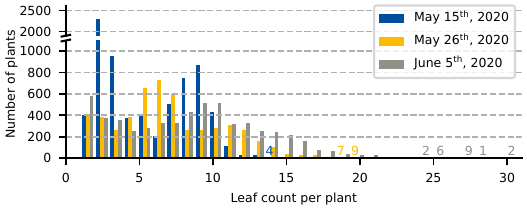}
    \caption{Distribution of leaf count of mostly visible plants in the training and validation set.}
    \label{fig:leaf_statistics}
  \end{figure}

Finally, we present in \cref{fig:leaf_statistics} the distribution of leaves per plant \change{per data collection day} of completely visible plants in the training and validation split.
\change{Similar to the trends for the canopy cover, we can also observe an increase in terms of the number of leaves over time. On May 20$^\text{th}$, most of the plants are still in the two-leaves stage with only a few plants in the later development with more than 10 leaves. Note that some leaves are also so-called germ leaves %
  that are later replaced by the real leaves. The peak in the leaf count shifts to the right on May 26$^\text{th}$ as the sugar beet plants develop more leaves in later growth stages. On the last data collection date, June 5$^\text{th}$, the distribution of leaves gets more long-tailed as now larger plants are competing for space. At this stage, however, it's also more likely that leaves are covered by other leaves, since we observe the field from a UAV.
  Thus, the true number of leaves is not observable.}

Overall, we annotated 583 unique crop plants at potentially different growth stages growing under real-world conditions in the open field.
Thus, the individual plant growth is affected by the weather conditions and the soil quality that changes over the whole field. As noted before, the visual appearance changes between different plants but also can have substantial differences due to the natural plant growth.
More specifically, 496 plants appear in all three dates, 15 plants in only two of the dates, and 72 plants only at a single point in time, which is caused by the conventional field management operations or natural growing conditions.

\section{Benchmarks}

In this section, we present the benchmark tasks that we provide together with the dataset.
These tasks cover different aspects of a perception system for the crop production domain in agriculture.
While we cover classical, well-established tasks, we also want to provide a novel task of hierarchical panoptic segmentation that provides a complete picture of the plant structure.

We provide metrics on the test set of our dataset including data from \emph{known} and \emph{unknown} fields for all investigated baseline approaches.
Note that we provide more details on the training setup, including hyperparameters and qualitative results, in the supplement.
We furthermore will provide code for the baselines in our code release.
In the supplement, we furthermore provide qualitative results together with more fine-grained quantitative results differentiating between the different fields of the test set.

\begin{table}
  \centering
  \small{
    \begin{tabular}{C{3cm}cccc}
      \toprule
      \multirow{2}{*}{Approach}       & \multirow{2}{*}{mIoU} & \multicolumn{3}{c}{IoU}                     \\
      \cmidrule(lr){3-5}
                                      &                       & Crop                    & Weed    & Soil    \\
      \midrule
      ERFNet~\cite{romera2017tits}    & $85.98$               & $94.30$                 & $64.37$ & $99.28$ \\ %
      DeepLabV3+~\cite{chen2017arxiv} & $85.97$               & $94.07$                 & $64.59$ & $99.25$ \\  %
      \bottomrule
    \end{tabular}
  }
  \caption{Baseline results for semantic segmentation on the \mbox{test set.}}
  \label{tab:semantic_segmentation}
\end{table}

\subsection{Semantic Segmentation}

\myparagraph{Task description.}
Semantic segmentation in images aims to train models capable of predicting each pixel's class.
Thus, we provide annotated ground truth data that assigns each pixel to the class soil, crop, or weed.
Consequently, an approach for this task needs to provide dense predictions assigning each pixel to one of the before-mentioned classes.

\myparagraph{State of the Art.}
Semantic segmentation is a classical task that was first mainly tackled using conditional random fields~\cite{ladicky2009iccv, kraehenbuehl2011nips} to exploit the neighboring structure of images.
With the advent of deep learning and the success in image classification~\cite{krizhevsky2012cacm}, dense prediction tasks are nowadays mainly tackled by encoder/decoder architectures~\cite{long2015cvpr-fcnf, romera2017tits, ronneberger2015micc}.
Recently, refined architectures add larger context~\cite{chen2018pami,chen2017arxiv} and multi-resolution processing~\cite{sun2019arxiv-hrfl} or rely on Transformers~\cite{vaswani2017neurips} for the encoder~\cite{xie2021neurips, cheng2021neurips}.
We refer to surveys~\cite{asgari2021air, lateef2019neuroc} for an overview of recent developments.

In the agricultural domain, most approaches~\cite{lottes2018ral,lottes2017jfr,milioto2018icra} follow the development and adopt the pipelines to account for the row structure~\cite{lottes2018ral} or
leverage additional background knowledge to cope with less labeled data~\cite{milioto2018icra}.

\myparagraph{Baselines. }
As baselines, we select DeepLabV3+~\cite{chen2017arxiv}\change{~(${39.8\text{ M}}$ params)} and ERFNet~\cite{romera2017tits}\change{~(${2.1\text{ M}}$~params)} at different ends of model capacity.%

\myparagraph{Metrics. }
To evaluate the performance of semantic segmentation models, we report the
common intersection-over-union~(IoU) for each class individually, where higher
values indicate a better performance~\cite{cordts2016cvpr}. Additionally, we
compute the mean intersection over union~(mIoU) across all classes as the main
metric.

\myparagraph{Results and Discussion.}
In \cref{tab:semantic_segmentation}, we show quantitative results of the
selected baselines. The investigated off-the-shelf semantic segmentation methods
already show an overall good performance in terms of mIoU. However, we observe a
relatively low IoU for weeds which are often wrongly assigned to pixels of
crops.
\change{We support these results qualitatively
in Fig.~10 
and~Fig.~11 of the supplement,
 depicting the
predictions of each approach as well as highlighting correct and false
predictions}. In terms of model capacity, the different investigated methods
perform very similarly, indicating that the models' capacity cannot resolve the
aforementioned issues. Surprisingly, the smaller, simpler, and faster
architecture ERFNet performs on par with the more complex DeepLabV3+ model that
commonly shows better performance in the context of autonomous driving.
\change{Furthermore, we refer to~ Tab.~9 
of the supplement for more detailed 
quantitative results distinguishing between each data collection date.}

\subsection{Panoptic Segmentation}

\myparagraph{Task description.} Panoptic segmentation~\cite{kirillov2019cvpr-ps} tackles the task of jointly estimating a pixel-wise semantic label and distinguishing instances.
This task differentiates between so-called ``stuff'' and ``thing'' classes.
The former corresponds to instance-less classes, \ie, soil, and the latter refers to classes with clearly separable objects, \ie, crops and weeds.
Consequently, an approach for this task needs to produce semantic masks assigning each pixel to crop, weed, or soil and an instance segmentation for crops and weeds.

\myparagraph{State of the Art.}
Most approaches for panoptic segmentation~\cite{kirillov2019cvpr-pfpn} extend classical semantic segmentation approaches with an instance branch or head to separate ``thing'' classes.
Generally, there are two main paradigms for generating instances prevalent: top-down and bottom-up approaches.
Top-down approaches~\cite{kirillov2019cvpr-pfpn, porzi2019cvpr, li2021cvpr-fcnf} use detection-based bounding box predictions to locate instances and mask predictions in bounding boxes to segment the located instances pioneered by Mask R-CNN~\cite{he2017iccv-mr}.
Bottom-up approaches~\cite{cheng2020cvpr, wang2020cvpr} use a separate decoder to estimate embedding vectors and offsets to find clusters corresponding to instances of ``thing'' classes guided by the semantic segmentation branch. %
The main focus of research in this field concentrates on improving the architecture to achieve better separation between instances~\cite{mohan2021ijcv, li2021cvpr-fcnf, porzi2021cvpr}.
However, recent approaches~\cite{cheng2022cvpr, yu2022cvpr, strudel2021iccv} based on Vision Transformer~\cite{dosovitskiy2021iclr} show substantial improvements.

In the agricultural domain, most methods adopt panoptic segmentation pipelines for crop and weed detection~\cite{champ2020apps, halstead2021fps}
to contribute towards sustainable crop production and targeted weed management in real field conditions.

\myparagraph{Baselines.}
We use Panoptic DeepLab~\cite{cheng2020cvpr}~\change{(${7.7\text{ M}}$ params)} and
Mask R-CNN~\cite{he2017iccv-mr}\change{~(${44.4\text{ M}}$ params)}.
Further, we show Mask2Former~\cite{cheng2022cvpr}\change{~(${44\text{ M}}$ params)}
performance of a Transformer-based approach.

\begin{table}
  \centering
  \small{
    \begin{tabular}{ccccc}
      \toprule
      Approach                              & PQ$^\dagger$ & $\text{PQ}_{\text{crop}}$ & $\text{PQ}_{\text{weed}}$ & $\text{IoU}_{\text{soil}}$ \\
      \midrule
      Panoptic DeepLab~\cite{cheng2020cvpr} & $57.97$      & $52.02$                   & $22.61$                   & $99.27$                    \\ %
      Mask R-CNN~\cite{he2017iccv-mr}       & $65.79$      & $67.61$                   & $31.30$                   & $98.47$                    \\ %
      Mask2Former~\cite{cheng2022cvpr}      & $69.99$      & $71.21$                   & $40.39$                   & $98.38$                    \\ %
      \bottomrule
    \end{tabular}
  }
  \caption{Baseline results for panoptic segmentation on the test set.}
  \label{tab:panoptic_segmentation}
  \vspace{0.08cm}
\end{table}

\myparagraph{Metrics.}
We separately compute the panoptic quality~\cite{kirillov2019cvpr-ps} for the
predicted instance masks of crops~($\text{PQ}_{\text{crop}}$) and
weeds~($\text{PQ}_{\text{weeds}}$). During evaluation, we treat predicted
instances associated with a partially visible instance, \ie, a plant where less
than $50$\% of its pixels are inside the image, as ``do not care'' regions not
affecting the score. Additionally, we report the IoU for the semantic
segmentation of soil~($\text{IoU}_{\text{soil}}$) to consider predictions
related to ``stuff''. In our final metric, we compute the average over
all three values and denote it as PQ$^\dagger$ as proposed by
Porzi~\etal~\cite{porzi2019cvpr}.

\myparagraph{Results and Discussion.}
In \cref{tab:panoptic_segmentation} we show that
Mask2Former~\cite{cheng2022cvpr} achieves the best overall performance.
\change{A more detailed quantitative evaluation provided 
in Tab.~11 
of the supplement, distinguishing between different data
collection days characterized by specific plant growth stages, reveals that the
instance segmentation of plants is challenging in cases of barely visible small
plants and large plants with high mutual overlap. We support these results
qualitatively in~Fig. 13 
and~Fig.~14 
of the supplement. 
This suggests that domain-specific models could potentially exploit the plant growth stage.}

\subsection{Detection}

\myparagraph{Task description.}
While pixel-wise segmentation of instances allows for extracting fine-grained information, often detecting instances is sufficient.
Therefore, we also propose using our data for studying plant or leaf detection in separate tasks.
For plant detection, we distinguish between the classes of crop and weed.
Similar to COCO~\cite{lin2014eccv}, we extract bounding box annotations from the instance-level plant and leaf annotations to allow training of object detection approaches.
An approach for either plant or leaf detection needs to provide bounding boxes and confidence scores for each detected instance.

\myparagraph{State of the Art.}
Early approaches for object detection relies on sliding window-based classification methods~\cite{viola2001ijcv} and research before 2014 mainly concentrates on better feature representations~\cite{dalal2005cvpr, gall2011pami}, part-based representations~\cite{felzenszwalb2010pami, leibe2004eccv}, or better proposal generation~\cite{vandersande2011iccv}.

Since 2013, CNN-based approaches have been prevalent as pioneered by R-CNN~\cite{girshick2014cvpr} and follow-up work~\cite{ren2015nips, girshick2015iccv, he2017iccv-mr}.
Generally, one can distinguish between single-stage and two-stage approaches.
Nowadays, single-stage approaches are mainly employed and YOLO~\cite{redmon2016cvpr}-based approaches are popular choices. %
Recently, also keypoint-based approaches~\cite{law2018eccv,zhou2019arxiv} were proposed that divert from the anchor-based methods.
Similarly to other tasks, the field recently shifted towards Transformer-based approaches~\cite{carion2020eccv}.%

In the agricultural domain, most methods use detectors to identify crops or weeds~\cite{halstead2021fps, saleem2022fps}
or suggest domain-specific adaptations, \eg, for fruit detection~\cite{mai2018icra}. %

\myparagraph{Baselines.}
We select established approaches for object detection, such as Faster
RCNN~\cite{ren2015nips}\change{~(${41.7\text{ M}}$ params)}, Mask R-CNN~\cite{he2017iccv-mr}\change{~(${44.4\text{ M}}$ params)} and
YOLOv7~\cite{wang2022arxiv}\change{~(${37.2\text{ M}}$ params)}, which are commonly used
approaches. Since this task refers to either plant or leaf detection, we train
models for each task separately. Although Mask R-CNN also provides an instance
segmentation, we do not consider these \change{here} but rely on its predicted bounding boxes.

\begin{table}
  \centering
  \tabcolsep3pt
  \small{
    \begin{tabular}{C{2.75cm}ccccc}
      \toprule
      \multirow{2}{*}{Approach}       & \multirow{2}{*}{mAP} & \multirow{2}{*}{mAP$_{50}$} & \multirow{2}{*}{mAP$_{75}$} & \multicolumn{2}{c}{AP}           \\
      \cmidrule(lr){5-6}
                                      &                      &                             &                             & Crop                   & Weed    \\
      \midrule
      Faster R-CNN~\cite{ren2015nips} & $40.43$              & $65.07$                     & $40.19$                     & $63.23$                & $17.62$ \\ %
      Mask R-CNN~\cite{he2017iccv-mr} & $38.68$              & $63.72$                     & $38.07$                     & $60.32$                & $17.05$ \\ %
      YOLOv7~\cite{wang2022arxiv}     & $60.48$              & $82.47$                     & $62.30$                     & $83.06$                & $37.91$ \\ %
      \bottomrule
    \end{tabular}
  }
  \caption{Baseline results for plant detection on the test set.}
  \label{tab:plant_detection}
\end{table}

\begin{table}
  \centering
  \small{
    \begin{tabular}{C{3cm}cccc}
      \toprule
      Approach                        & mAP     & mAP$_{50}$ & mAP$_{75}$ \\
      \midrule
      Faster R-CNN~\cite{ren2015nips} & $33.91$ & $64.61$    & $31.30$    \\ %
      Mask R-CNN~\cite{he2017iccv-mr} & $34.41$ & $66.02$    & $32.15$    \\ %
      YOLOv7~\cite{wang2022arxiv}     & $57.90$ & $86.85$    & $62.92$    \\ %
      \bottomrule
    \end{tabular}
  }
  \caption{Baseline results for leaf detection on the test set.}
  \label{tab:leaf_detection}
\end{table}

\myparagraph{Metrics.}
In line with established benchmarks~\cite{everingham2010ijcv,everingham2015ijcv,lin2014eccv}, we
report the average precision~(AP) for each class and mean average precision~(mAP) across all classes, which uses multiple IoUs for matching between $0.5$ and
$0.95$ with a step size of $0.05$. Furthermore, we report the mean average
precision at $0.5$~IoU (mAP$_{50}$) and $0.75$~IoU (mAP$_{75}$). As
previously, we treat each predicted bounding box associated with a partially
visible instance as ``do not care'' regions. Thus, these predictions do not affect the scores.

\myparagraph{Results and Discussion.}
In \cref{tab:plant_detection}, we show results for plant detection, where we see
that modern approaches have a clear edge over the other approaches.
Apparently, weed detection is more difficult than crop detection, which could
result from smaller plant sizes\change{, as also suggested qualitatively
in~Fig.~16 and Fig.~17 
of the supplement.}

In
\cref{tab:leaf_detection}, we summarize the results for leaf detection, which
shows lower performance across all methods compared with aforementioned plant detection,
indicating the need for domain-specific approaches.
\change{In~Tab.~15 
of the supplement, we provide more detailed results
for each data collection day and additionally show qualitative
results in~Fig.~18 and Fig. 19 
of the supplement.}

\subsection{Leaf Instance Segmentation}

\myparagraph{Task description.}
Leaf instance segmentation is relevant for estimating the growth stage of a plant~\cite{lancashire1991aab} and also the basis for leaf disease detection~\cite{mohanty2016fps}.
Such approaches are involved in phenotyping activities to investigate new varieties of crops~\cite{minervini2015spm}.
An automatic, {vision-based} assessment of such traits has the potential to have reproducible and objective measurements at a high temporal frequency.
Consequently, an approach for this task needs to predict an instance mask for each visible crop leaf.

\begin{table}[t]
  \centering
  \small{
    \begin{tabular}{cC{3cm}}
      \toprule
      Approach                         & $\text{PQ}_{\text{leaf}}$ \\
      \midrule
      Mask R-CNN~\cite{he2017iccv-mr}  & $59.74$                   \\ %
      Mask2Former~\cite{cheng2022cvpr} & $57.50$                   \\ %
      \bottomrule
    \end{tabular}
  }
  \caption{Baseline results for leaf instance segmentation on test set.}
  \label{tab:leaf_instance_segmentation}
\end{table}

\myparagraph{State of the Art.}
Instance segmentation is closely related to object detection. Therefore earlier approaches rely on object detection approaches~\cite{ren2015nips,redmon2016cvpr} to perform top-down instance segmentation by predicting segmentation masks for bounding boxes~\cite{he2017iccv-mr, bolya2022pami}.
A different line of research~\cite{debrabandere2017cvprws} investigated the usage of bottom-up processing, where first pixel-wise embedding vectors are estimated such that pixels belonging to the same instance are near in embedding space, while embedding vectors of different instances are separated. The estimated embedding vectors can then be clustered, resulting in instances.
Recently, several methods~\cite{wang2020eccv, wang2020neurips} were proposed that directly estimate masks for each object instance.
Most recently, also Transformer-based approaches~\cite{lazarow2022cvpr, cheng2022cvpr} for instance segmentation gained interest.
Popularized by CVPPP~\cite{minervini2016prl}, several approaches tackle the task of leaf instance segmentation~\cite{huang2022aaai} or leaf counting~\cite{weyler2021ral}.

\myparagraph{Baselines.}
As baselines for our experiments, we employ \text{Mask R-CNN}~\cite{he2017iccv-mr}\change{~(${44.4\text{ M}}$ params)} and
Mask2Former~\cite{cheng2022cvpr}\change{~(${44\text{ M}}$ params)}. While the former method represents a
traditional top-down approach, the latter belongs to more recent
methods relying on a Transformer decoder and masked attention.

\myparagraph{Metrics.}
We compute the panoptic quality~\cite{kirillov2019cvpr-ps} for the predicted instance masks of crop leaves, denoted as $\text{PQ}_{\text{leaf}}$.
As previously, any instance prediction associated with a partially visible instance does not affect the score.

\myparagraph{Results and Discussion.}
\cref{tab:leaf_instance_segmentation} shows the results of the investigated
baselines. In this setting, the approaches generally struggle to separate leaves\change{, as they are naturally overlapping, even for smaller plants.
In~Fig.~20 
and~Fig.~21 
of the supplement,
we support these results qualitatively and provide more detailed metrics differentiating
between each data collection day in~Tab.~17~
of the supplement.
}
Again, we suspect that more domain-specific approaches
could induce prior knowledge to achieve a better separation.

\subsection{Hierarchical Panoptic Segmentation}
\label{sec:hierarchical_panoptic_segmentation_benchmark}

\myparagraph{Task description.}
Models for hierarchical panoptic segmentation target objects, which can be
represented as an aggregation of individual parts, \eg, plants can be
represented as the union of their leaves~\cite{weyler2022wacv}. Consequently,
these methods provide a simultaneous instance segmentation of the whole object
and each part. Thus, they are capable of providing more detailed information
about each object, \eg, the association of individual leaves to a specific plant
allows obtaining the total number of leaves per plant, which correlates to its
growth stage~\cite{lancashire1991aab}. We provide the annotated instance masks
of all crops and their associated leaves. Since there are no leaf annotations
for weeds, we do not consider them under the guise of a hierarchical structure.
Thus, we also relate to weeds as ``stuff'' for this task.

\begin{table}
  \centering
  \tabcolsep3pt
  \small{
    \begin{tabular}{ccccccc}
      \toprule
      \multirow{2}{*}{Approach}       & \multirow{2}{*}{PQ$^\dagger$} & \multirow{2}{*}{PQ} & \multirow{2}{*}{$\text{PQ}_{\text{crop}}$} & \multirow{2}{*}{$\text{PQ}_{\text{leaf}}$} & \multicolumn{2}{c}{IoU}           \\
      \cmidrule(lr){6-7}              &                               &                     &                                            &                                            & Weed                    & Soil    \\
      \midrule
      HAPT~\cite{roggiolani2023icra} & $65.27$                       & $50.73$             & $54.61$                                    & $46.84$                                    & $61.11$                 & $98.50$ \\ %
      Weyler \etal~\cite{weyler2022wacv}    & -                             & $40.49$             & $38.37$                                    & $42.60$                                    & -                       & -       \\ %
      \bottomrule
    \end{tabular}
  }
  \caption{Baseline results for hierarchical panoptic segmentation on the test set.}
  \label{tab:hierarchical_panoptic_segmentation}
  \vspace{0.2cm}
\end{table}

\myparagraph{State of the Art.}
Several recent works exploit the underlying hierarchical structure of objects
to obtain a panoptic segmentation~\cite{roggiolani2023icra, weyler2022wacv}.
In the agricultural domain, recent methods~\cite{roggiolani2023icra, weyler2022wacv}
operating in real field conditions exploit the hierarchical structure of plants
to predict the instance segmentation of individual crops and their leaves.

\myparagraph{Baselines.}
We select the methods by Weyler~\etal~\cite{weyler2022wacv}\change{~(${2.2\text{
M}}$ params)} and
Roggiolani~\etal~\cite{roggiolani2023icra}\change{~(${2.4\text{ M}}$ params)}
as baselines that both perform a simultaneous instance segmentation of crops and their
associated leaves, \change{where the latter method is denoted as HAPT}. The first method is a bottom-up approach that first predicts
leaves, which are then associated to a plant. In contrast, HAPT uses a
hierarchical feature aggregation starting at the plants providing plant-level
features to then predict leaves.

\myparagraph{Metrics.}
To evaluate the performance of this task, we compute the panoptic quality~\cite{kirillov2019cvpr-ps} for the predicted instance masks of all crops ($\text{PQ}_{\text{crop}}$)
and leaves ($\text{PQ}_{\text{leaf}}$) separately. We report the average panoptic quality over both values, denoted as PQ.
As previously, any instance prediction assigned to a partially visible instance does not affect the metrics.
To account for methods that filter pixels related to weeds or soil with an additional semantic segmentation, we also report
the IoU for both classes. Finally, we compute PQ$^\dagger$ as the average over $\text{PQ}_{\text{crop}}$, $\text{PQ}_{\text{leaf}}$,
and both IoU values.

\myparagraph{Results and Discussion.}
In~\cref{tab:hierarchical_panoptic_segmentation}, we show the results of the
hierarchical approaches. Here, we can see that both methods do not obtain
consistent predictions for plants at a large growth stage, where individual
plants and their leaves overlap. In particular, instance separation of leaves
seems most challenging in line with the plant instance segmentation. Thus,
methods targeting these scenarios could improve the performance. \change{We
support these findings in~Fig.~19 
of the supplement,
where we perform the evaluation for each data collection day separately.
Ultimately, we show quantitative results
in~Fig.~22
, which we separate into true
positives, false positives, and false negatives
in~Fig.~23 
in the supplement.}

\change{\section{Challenge in Conjunction with CVPPA Workshop at IEEE/CVF ICCV 2023}}

\change{In conjunction with the workshop on Computer Vision in Plant Phenotyping and Agriculture held at the IEEE/CVF International Conference on Computer Vision (ICCV) in 2023, we invited the community to tackle the most challenging task of hierarchical panoptic segmentation using our dataset.
  We received overall 148 submissions from 107 registered participants on the competition hosted on CodaLab\footnote{The concluded and now closed competition is still available at \url{https://codalab.lisn.upsaclay.fr/competitions/13904}.}, where one could upload predictions until a fixed deadline.}

\change{For the top-performing entries of the leaderboard, we invited authors to provide a technical report of their approach\footnote{Non-archival, non-peer reviewed technical reports are available at \url{https://cvppa2023.github.io/challenges/}}. The technical solutions surpassed the baselines by a large margin and often employed the Segment Anything Model~\cite{kirillov2023iccv} either in conjunction with a detection approach or initial segmentation that is refined.
But also a Mask2Former-based~\cite{cheng2022cvpr} approach using a mask refinement on small plants and a second stage for leaf instance segmentation on plant masks showed promising results surpassing our off-the-shelf baselines presented in \cref{sec:hierarchical_panoptic_segmentation_benchmark}.
}

\section{Potential Impact on Other Topics}

Besides the already covered supervised tasks in agricultural perception, our dataset providing labeled and unlabeled images  has the potential to impact also other fields of research and applications in the agricultural domain, such as research in self-supervised representation learning, domain generalization, and unsupervised domain adaptation that is currently getting increasing interest in the computer vision and robotics community. Exploiting developments in semi-supervised, but also unsupervised learning of vision models seems like a indispensable step to reduce the burden of annotating data and unlocking the scalable deployment of vision models in the agricultural domain.

Furthermore, the combination with other agricultural datasets providing pixel-wise annotations, \eg, GrowliFlowers~\cite{kierdorf2022jfr}, opens the door for studying cross-domain transfer between different plant species towards the goal of developing more generalizable visual perception systems in the agricultural domain.

\section{Conclusion}

In this paper, we present a novel dataset for studying visual perception in the agricultural domain of crop production using real-world field images captured by an UAV.
Together with dense pixel-wise annotations of crops and weeds that distinguish instances of plants, we also provide leaf-level pixel-wise annotations of crop leaves.

In line with the dataset, we presented our benchmark tasks that will be evaluated on a hidden test set to allow an unbiased and controlled evaluation of developed approaches.
The server-side evaluation also ensures that metrics are consistent and reliable allowing to compare approaches based on published results.

For each task, we also provide baseline results that show the performance of off-the-shelf approaches for the different tasks.
These results show that certain tasks need further research to tackle the specific challenges of the agricultural domain.
We believe that more domain-specific approaches exploiting domain knowledge could boost performance.

\appendices

\ifCLASSOPTIONcompsoc
  \section*{Acknowledgments}
\else
  \section*{Acknowledgment}
\fi
We thank all students annotating the data.
The work has been funded by~the  Deutsche Forschungsgemeinschaft (DFG, German~Research~Foundation)
under Germany’s Excellence Strategy, EXC-2070 -- 390732324 (PhenoRob).

\bibliographystyle{IEEEtranS}
\bibliography{glorified, new}

\begin{IEEEbiography}[{\vspace{-1.2cm}\includegraphics[width=1in,height=1.25in,clip,keepaspectratio]{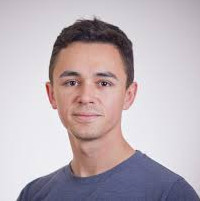}}]{Jan Weyler} is a PhD student in Engineering at the Photogrammetry \& Robotics Lab at the University of Bonn, Germany. He obtained his B.Sc. in 2015 and his M.Sc. degree in Geodesy and Geoinformation in 2019 from the University of Bonn, Germany. His research focuses on vision-based semantic scene understanding for agricultural robots.\vspace{-1.2cm}
\end{IEEEbiography}
\begin{IEEEbiography}[{\includegraphics[width=1in,height=1.25in,clip,keepaspectratio]{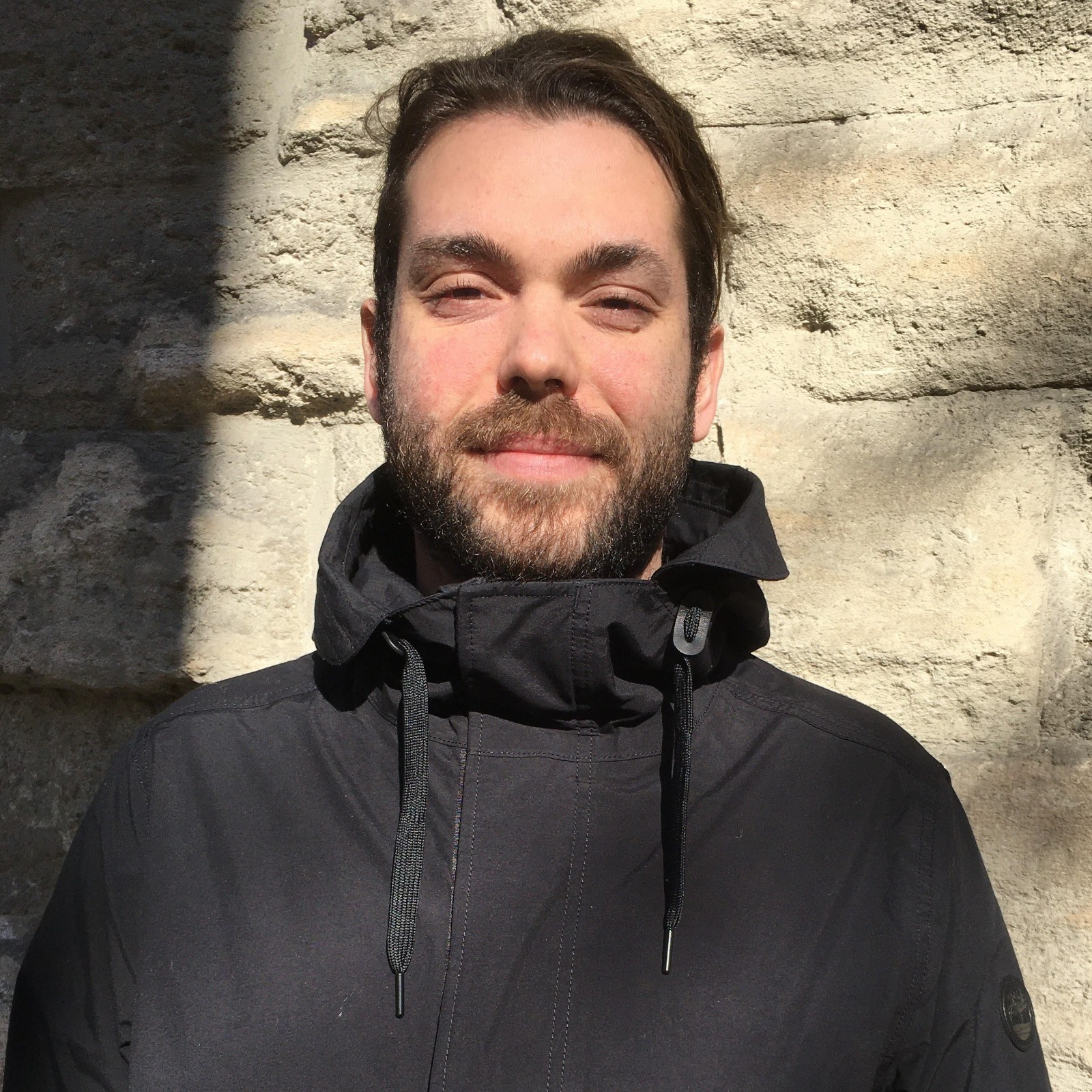}}]{Federico Magistri} is a Ph.D. student at the Photogrammetry \& Robotics Lab at the University of Bonn, Germany, since November 2019. He received his M.Sc. in Artificial Intelligence and Robotics from ''La Sapienza'' University of Rome, Italy, with a thesis on Swarm Robotics for Precision Agriculture in collaboration with the National Research Council of Italy and the Wageningen University and Research, Netherlands.\vspace{-1cm} %
\end{IEEEbiography}
\begin{IEEEbiography}[{\includegraphics[width=1in,height=1.25in,clip,keepaspectratio]{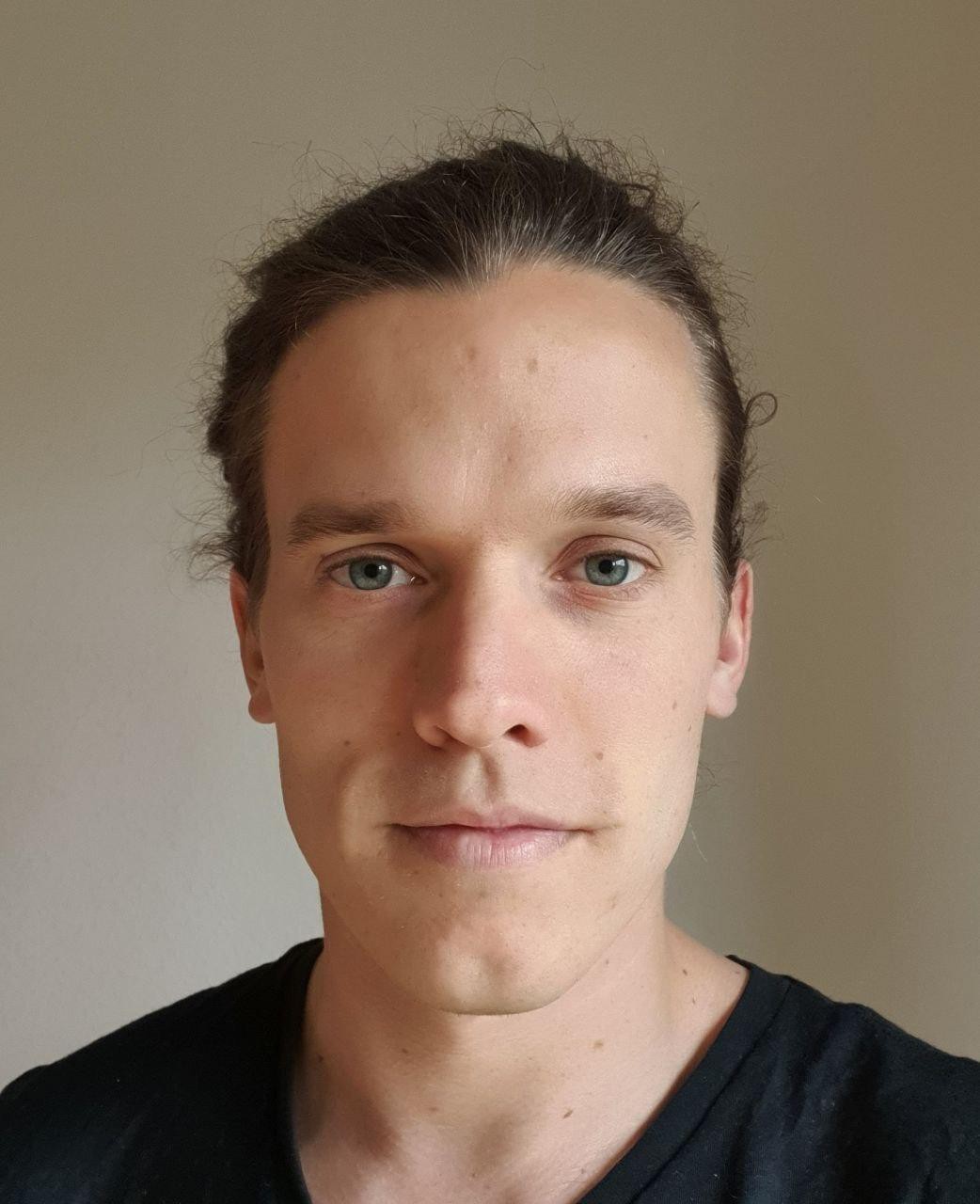}}]{Elias Marks} is a PhD student in Engineering at the Photogrammetry \& Robotics Lab at the University of Bonn, Germany. He obtained his B.Sc. degree in Robotics and Automation from the Hochschule Heilbronn, Germany, in 2018 and received his M.Sc. degree in Artificial Intelligence and Robotics at University La Sapienza in Rome, Italy, in 2021. His research focuses on plant modeling for phenotyping based on image data.\vspace{-1cm}
\end{IEEEbiography}
\begin{IEEEbiography}[{\includegraphics[width=1in,height=1.25in,clip,keepaspectratio]{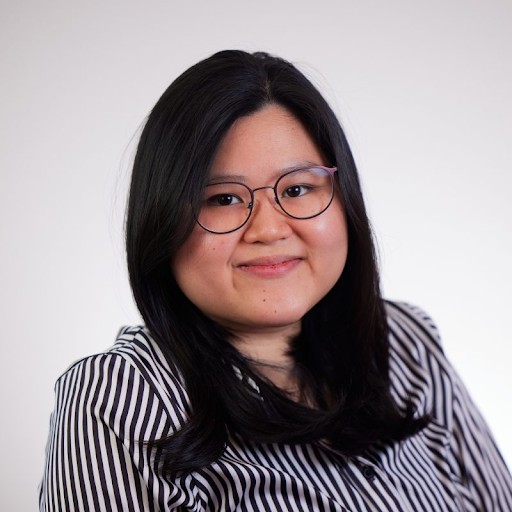}}]{Yue Linn Chong} is a Ph.D. student in Engineering at Photogrammetry \& Robotics Lab at the University of Bonn, Germany.
  She completed her B.Eng in Mechanical Engineering from the National University of Singapore in 2017.
  In 2020, she completed her M.Sc. in Mechanical Engineering from the National University of Singapore.
  Her research focuses on unsupervised learning using generative models.\vspace{-1cm}
\end{IEEEbiography}
\begin{IEEEbiography}[{\includegraphics[width=1in,height=1.25in,clip,keepaspectratio]{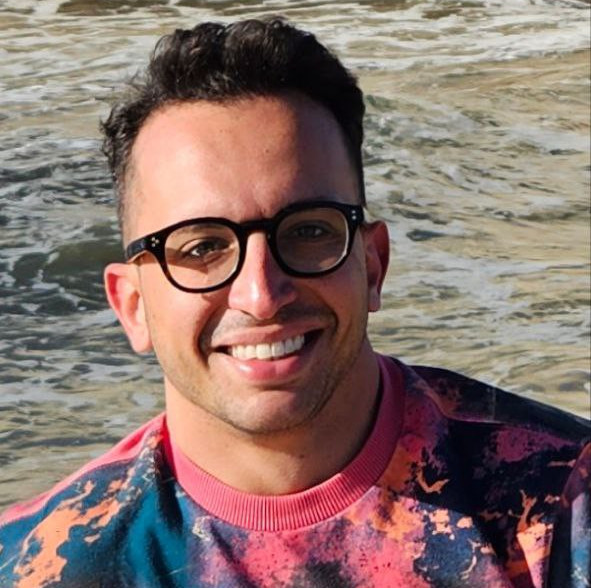}}]{Matteo Sodano} is a PhD student in Engineering at the Photogrammetry \& Robotics Lab at the University of Bonn since January 2021. He obtained his MSc degree in Control Engineering in 2020. His research centers around perception and segmentation, with a focus on novel object discovery.\vspace{-1cm}
\end{IEEEbiography}
\begin{IEEEbiography}[{\includegraphics[width=1in,height=1.25in,clip,keepaspectratio]{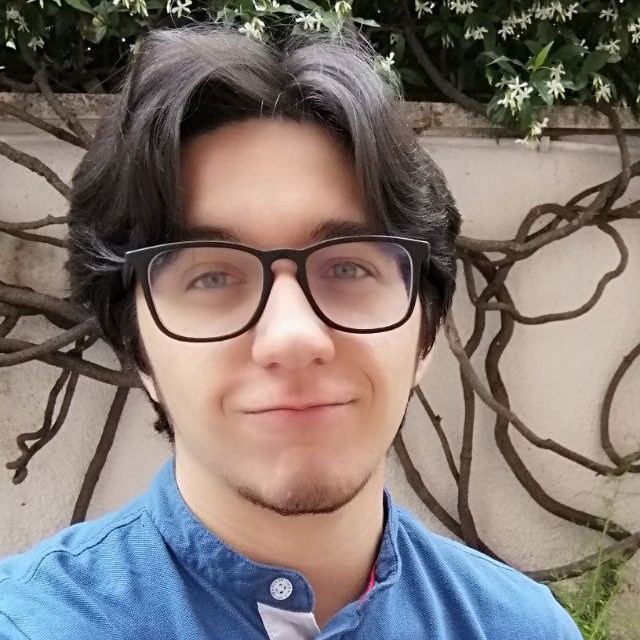}}]{Gianmarco Roggiolani} is a Ph.D. candidate in the Photogrammetry \& Robotics Lab at the University of Bonn, Germany. He obtained his B.Sc. degree in Computer and Automatic Engineering in 2018 and received his MSc degree in Artificial Intelligence and Robotics in 2021, both from the Sapienza University of Rome, Italy. His research focuses on self-supervised techniques to improve the performance of vision-based learning systems in agricultural robotics. \vspace{-1cm}
\end{IEEEbiography}
\begin{IEEEbiography}[{\includegraphics[width=1in,height=1.25in,clip,keepaspectratio]{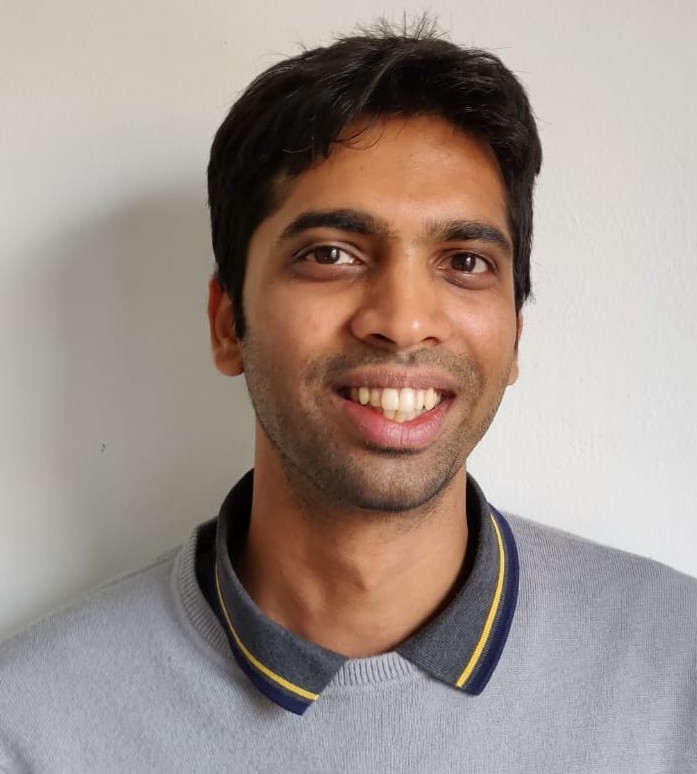}}]{Nived Chebrolu} is a postdoctoral research associate at the Oxford Robotics Institute, University of Oxford, UK. His research interests are in developing robust localization and mapping techniques for field robotics applications. He obtained his Ph.D. from the University of Bonn in 2021, where he developed registration techniques for agricultural robotic applications. Before that, Nived received his M.Sc. in Robotics from Ecole Centrale de Nantes~(ECN), France, and the University of Genoa, Italy in 2015.\vspace{-1cm}
\end{IEEEbiography}
\begin{IEEEbiography}[{\includegraphics[width=1in,height=1.25in,clip,keepaspectratio]{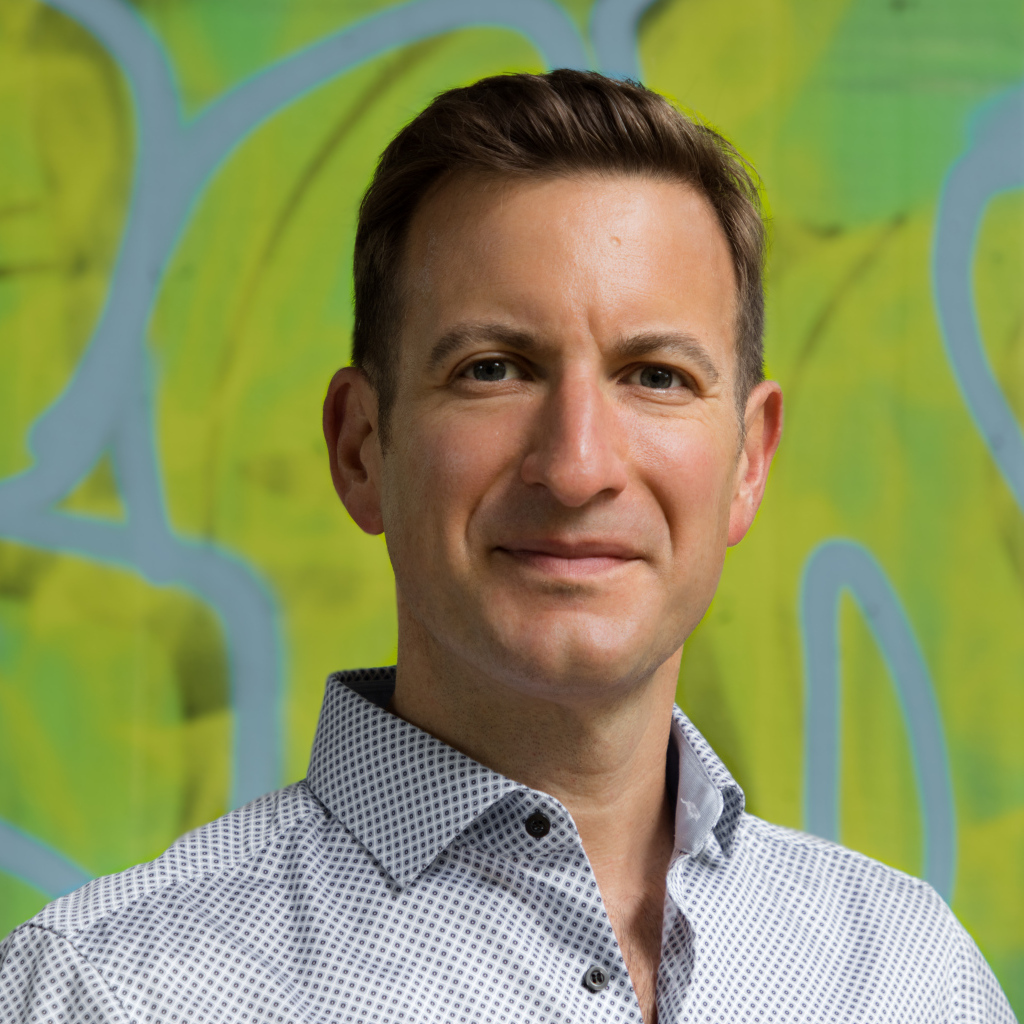}}]{Cyrill Stachniss} is a full professor at the University of Bonn, Germany, with the University of Oxford, UK, as well as with the Lamarr Institute for Machine Learning and AI, Germany. He is the Spokesperson of the DFG Cluster of Excellence PhenoRob at the University of Bonn. His research focuses on probabilistic techniques and learning approaches for mobile robotics, perception, and navigation. Main application areas of his research are agricultural and service robotics and self-driving cars.\vspace{-1cm}
\end{IEEEbiography}
\begin{IEEEbiography}[{\includegraphics[width=1in,height=1.25in,clip,keepaspectratio]{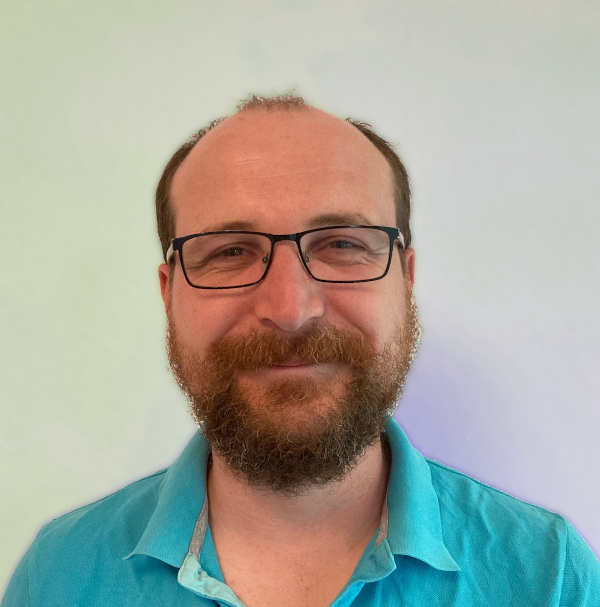}}]{Jens Behley} received his Dipl.-Inform.  in computer science in 2009 and his Ph.D. in computer science in  2014, both from the Dept. of Computer Science at the University of Bonn, Germany. Since 2016, he is a postdoctoral researcher at the Photogrammetry \& Robotics Lab at the University of Bonn, Germany. He finished his habilitation at the University of Bonn in 2023. His area of interest lies in the area of perception for autonomous vehicles, deep learning for semantic interpretation, and LiDAR-based SLAM.\vspace{-1cm}
\end{IEEEbiography}

\end{document}